\documentclass[accepted]{uai2026} 
                        
\usepackage{macro}

\usepackage{hyperref}       
\usepackage{url}            
\usepackage{booktabs}       
\usepackage{amsfonts}       
\usepackage{nicefrac}       
\usepackage{microtype}      
\usepackage{xcolor}         %

\usepackage[american]{babel}

\usepackage{natbib} 
\usepackage{mathtools} 
\usepackage{booktabs} 
\usepackage{tikz} 

\title{Meta-Dependence in Conditional Independence Testing}

\author[1,2]{\href{mailto:<bijan@dartmouth.edu>?Subject=Meta-Dependence in Conditional Independence Testing}{Bijan H. S. Mazaheri}{}}
\author[2, 3]{Jiaqi Zhang}
\author[2, 3]{Caroline Uhler}
\affil[1]{%
    Thayer School of Engineering\\
    Dartmouth College\\
    Hanover, New Hampshire, USA
}
\affil[2]{%
    Schmidt Center\\
    Broad Institute of MIT and Harvard\\
    Cambridge, Massachusetts, USA
}
\affil[3]{%
    Laboratory for Information and Decision Systems\\
    Massachusetts Institute of Technology\\
    Cambridge, Massachusetts, USA
  }
  
\begin{document}
\maketitle

\begin{abstract}
  Conditional independence testing is a critical component of feature screening, invariant statistical models, and causal discovery. Many of these algorithms rely on the sequential application of conditional independence tests, and their stability hinges on how their outcomes interact. 
We study this ``meta-dependence'' between conditional independence properties using the following geometric intuition:
satisfying each conditional independence property constrains the space of possible joint distributions to a manifold.
The ``meta-dependence'' of multiple conditional independences in a probability distribution is informed by its position relative to these manifolds. 
We provide a simple-to-compute measure of this meta-dependence using moment projections, with a closed-form expression for multivariate Gaussian distributions, and consolidate our findings empirically using both synthetic and real-world data. Our measure of meta-dependence does not rely on graphical properties of the distribution and can be computed directly from summary statistics such as a covariance matrix, allowing for various applications. We demonstrate one use case of meta-dependence, using a simple redundancy metric to tune significance thresholds and improve causal discovery.
\end{abstract}

\section{INTRODUCTION}\label{sec:intro}
Structural Causal Models (SCMs), popularized by \citet{pearl1998graphs}, are networks of causal dependencies that drive data-generating processes. Knowing the graphical structure of these networks enables us to identify causal effects \citep{pearl2009causality}, efficiently search for root causes \citep{ikram_root_2022}, and design more informative experiments \citep{zhang_active_2023}. In practice, however, the SCM is rarely available a priori and must be inferred from data. This is known as \emph{causal discovery} \citep{spirtes2000causation}.

Data generated from SCMs exhibit predictable conditional independence (CI) structures from the causal Markov condition, and these structures can be read off the causal graph. For example, a chain $A \rightarrow B \rightarrow C$ exhibits a Markov property whereby the first and last random variables are independent given the middle one. \emph{D-separation} conditions formalize exactly how graphical structure constrains the joint distribution with respect to conditional independence, and the resulting properties are the machinery underlying most graphical methods for causal identification \citep{pearl2009causality}. The same reasoning underlies methods well beyond effect identification, from invariant feature selection \citep{rojas2018invariant} to the transportability of causal and statistical findings across settings \citep{pearl_external_2014}.

\paragraph{Motivation} Conditional independence properties are not separate degrees of freedom within a causal system. The ``graphoid axioms'' \citep{pearl_graphoids_1986} and the ``Verma constraints'' \citep{verma1990causal} (the latter arising with latent variables) capture how CIs constrain one another, but these are \emph{qualitative, graph-level} rules. Such dependence matters for any algorithm that uses multiple CI tests: if an algorithm fails when \emph{at least one test} is incorrect, independent failures are costlier than co-occurring ones. This is most stark in hypothesis testing, where the standard Bonferroni correction treats test errors as independent, overcorrecting when they are redundant. Yet no metric quantifies the information shared between CI tests at the level of the distribution, let alone one computable without the causal graph. We provide a geometric insight into this dependence and a simple-to-compute metric for it.

\subsection{Problem and Summary}\label{sec:contri}
Suppose we repeatedly subsample a dataset, keeping track of the outcomes of two CI tests. Under this stochastic process, the two tests become random variables that can correlate. We call this phenomenon ``meta-dependence,'' and we call its sub-sampling estimate ``Finite Sample CI Dependence'' (FS-CID). While FS-CID can clearly be computed by the bootstrapped approach just described, it is time-consuming and provides no mathematical insight into the reason for these dependencies.

In this paper, we quantify the meta-dependence between different CI tests geometrically. We do so without access to a causal graph or the raw data, requiring only summary statistics of the distribution (e.g., a covariance matrix). The geometric picture comes from intuition built in Section~\ref{sec: CI Dependence Examples}, which shows that a slight change to a distribution may bring it closer to some CI properties and farther from others.

Such distribution ``changes'' are easy to interpret as perturbations on the parameters of the structural equations, but this framework disappears when the parametric form of the underlying data-generating distribution is unknown. As such, Section~\ref{sec: definition} extends the intuition to a ``moment projection'' \citep{nielsen2018information} from an empirical distribution onto a manifold of distributions that obey some CI property. We use these moment projections to formally define Conditional Independence Meta-Dependence (CIMD), which measures the change in Kullback--Leibler (KL) divergence associated with one CI property when the distribution is projected onto another. To make these notions practical, Section~\ref{sec: empirica} shows that both moment projections and CIMD have closed forms for multivariate Gaussian distributions.

We find that CIMD can be positive or negative and, somewhat surprisingly, that such variations can arise within models that obey the \emph{same causal structure} --- only differing in their structural equations. These observations suggest that the co-occurrence or anti-correlation of CI-testing errors is not a purely graphical phenomenon (as is the case for graphoid axioms and Verma constraints), but also driven by the underlying ground-truth probability distribution. This dependence manifests in finite-sample dependence as follows: finite-sample uncertainty perturbs the ground-truth distribution into an empirical one that is likely ``close'' in distributional distance,\footnote{We will use the KL divergence, which is not a distance, but is close in concept.} so properties that correspond to similar perturbations are likely to co-occur. We establish this link by showing that variations in CI-test outcomes under FS-CID simulations closely match the purely distributional CIMD for both synthetic and real-world data (Section~\ref{sec: empirical}).

While the implications of this insight are fundamental, our results also have immediate consequences for the tuning of algorithms that rely on CI tests. To support this claim, we provide a simple adjustment to the widely used PC algorithm from causal discovery that assigns different significance thresholds across tests (Section~\ref{sec: causal discovery}), improving accuracy in data-sparse settings.

\subsection{Related Works}\label{app:motiv-rela}

\paragraph{Causal Discovery.} Causal discovery is the task of recovering a causal structure from data, often in the form of a \emph{directed acyclic graph} (DAG) or a representation of an equivalence class of DAGs (see \citet{vowels_dya_2022, squires2023causal} for reviews).
One approach to causal discovery involves a \emph{constraint-based} search guided by \emph{many} CI tests, e.g., the PC-algorithm from \citet{spirtes2000causation}. As such, meta-dependence between CI properties is of great interest to researchers in this area, both for tuning of hypothesis significance thresholds and in understanding the problem's inherent complexity.

\paragraph{Causal Strength and Stability.} Related work quantifies the strength of \emph{causal} relationships \citep{janzing2013quantifying} and the stability of causal models \citep{schulman2016stability}. These notions analyze causal quantities and assume a \emph{known} structure; we instead target causal discovery, where the structure is \emph{unknown}.

\paragraph{Complexity of DAGs.} Relatedly, some DAGs are ``harder'' to learn than others --- a ``DAG complexity'' tied to the data needed to verify a structure. One such quantity is the number of CI queries required to confirm a structure \citep{zhang2024membership}, but this purely graphical count ignores the distribution-driven \emph{dependencies} between queries that we study here. A related notion is ``self-compatibility'' with respect to structures on subsets of variables \citep{faller2024self}. Self-compatibility posits that changes in the recovered structure on different subsets of the observed variables contain information about the stability of the resulting graphical properties. Self-compatibility is likely related to the information geometry of the empirical distribution, since the exclusion of some observables removes potentially redundant CI tests from constraint-based algorithms.

\paragraph{Feature Selection.}
Many feature-selection algorithms take advantage of the observation that conditioning on a sufficiently informative set of features makes the rest of the features independent from the label that we seek to predict \citep{tsamardinos2003towards}. Such a set has been characterized as a Markov blanket or Markov boundary \citep{richardson2003markov, pearl2009causality} (see
\citet{yu2020causality} for a review of causality-informed feature selection procedures).
Causal structure has also been applied to understand relationship invariances in the presence of shifting domains and distributions \citep{magliacane2018domain, rojas2018invariant, mazaheri2023causal}, and these domains may benefit from a  better understanding of CI meta-dependence.

\section{PRELIMINARIES}\label{sec:prelim}
\paragraph{Notation.}
We will generally use the capital Roman alphabet to denote random variables (e.g., $A, B, C, V$) and the lowercase Roman alphabet to denote assignments to those random variables (e.g., $A=a$ or just $a$). We will use blackboard bold font to denote the set of possible values a variable can take, e.g., $a \in \mathbb{A}$. Bold will indicate a set of random variables, e.g., $\bvec{V} = \{V_1, V_2, \ldots \}$, and $\bvec{v}$ is an assignment to $\bvec{V}$. Parents and children in graphs will also follow these conventions, e.g., $\PA(V)=\pa^{\bvec{v}}(V)$, where the assignments to those parents come from values specified in $\bvec{v}$. We sometimes use subscripts to indicate the relevant graph structure for the parents, e.g., $\PA_{\mathcal{G}}(V)$, but this may be omitted without ambiguity. 
We will generally use the Greek alphabet (e.g., $\alpha, \beta$) to represent parameters for structural equations.

\paragraph{Structural Causal Models.}
A structural causal model (SCM) is a tuple $\langle \mathcal{G}, P\rangle$ consisting of a directed acyclic graph (DAG) $\mathcal{G}$ and a probability measure on the variables (which are vertices of $\mathcal{G}$) \citep{pearl2009causality}. The distribution factorizes according to $\mathcal{G}$:
\begin{equation}
    P(\bvec{v}) = \prod_{V \in \bvec{V}} P(v \given \pa^{\bvec{v}}_{\mathcal{G}}(v)).
\end{equation}
This gives rise to other conditional independencies from the causal Markov condition that can be described using the d-separation rules. If two variables $A, B$ are d-separated by $\bvec{C}$, we write $A \dsep B \given \bvec{C}$. The causal Markov condition gives $A \dsep B \given \bvec{C} \Rightarrow A \independent B \given \bvec{C}$.

\paragraph{Faithfulness.}
The formulation of structural causal models yeilds the causal Markov condition, but the converse is not necessarily true \citep{ramsey2012adjacency}. That is, an ``unblocked'' causal pathway between two variables does not necessarily require that they be statistically dependent. Statistical dependence from causal linkage is called ``faithfulness,'' which is often assumed in order to give a two-way correspondence between CI-test outcomes and graphical properties.
The success of constraint-based causal discovery hinges on this correspondence between graphical properties (d-separation) and statistical independence.

\paragraph{Strong Faithfulness.}
Perfect statistical independence is extremely unlikely within empirical probability distributions. To address this, \citet{zhang2012strong} introduced a stronger notion of faithfulness that requires a minimal threshold of dependence between all causally linked variables; a closely related assumption underlies the high-dimensional analysis of the PC-algorithm by \citet{kalisch_estimating_2007}. For mathematical simplicity, the definition is only given for linear structural equations with additive Gaussian noise, where conditional dependence can be quantified using partial correlation.
\begin{definition}[$\lambda$-strong faithfulness]\label{def:strong-faithfulness}
    For constant $\lambda\in (0,1)$, a multivariate Gaussian distribution $P$ is said to be $\lambda$-strong-faithful to DAG $\mathcal{G}$ \emph{with respect to} two variables $A,B$ and set $\bvec{C}$,\footnote{Here $\bvec{C}\cap\{A,B\}=\varnothing$. In cases where $\bvec{C}=\varnothing$, we omit it for simplicity.} if and only if the absolute value of the partial correlation between $A,B$ conditioned on $\bvec{C}$ (denoted $\rho_{A,B\mid \bvec{C}}$) is bounded below by $\lambda$. That is, $\lambda$-strong faithfulness requires
    \begin{equation*}
        A \not\dsep B \given \bvec{C}\quad\Longleftrightarrow\quad|\rho_{A,B\mid \bvec{C}}|>\lambda.
    \end{equation*}
\end{definition}

The introduction of $\lambda$-strong faithfulness enabled a more critical examination into the faithfulness assumption. While a set of unfaithful distributions has Lebesgue measure $0$, the study of \citet{uhler2013geometry} showed that the geometry of unfaithful distributions is a sufficiently ``bendy'' manifold such that the majority of distributions are ``close'' to an unfaithful distribution. As such, most models will be in danger of (empirically) deviating towards a faithfulness violation, leading to unstable results. This paper utilizes this geometric perspective to study how these instabilities are related, which serves as a fundamental motivation for the perspective we pose on CI meta-dependencies.

\paragraph{Information-theoretic Measures.}
While strong faithfulness was originally defined with respect to Gaussian vectors with linear dependencies, we will generalize this notion using information theory. Kullback--Leibler (KL) divergence gives an asymmetric notion of difference between distributions \citep{kullback1997information}. For two distributions $P$ and $Q$ over $x \in \mathbb{X}$ with densities (or, for discrete variables, probability mass functions) $p$ and $q$, the KL divergence is the expected log-likelihood ratio under $P$:
\begin{equation}
    D(P || Q) \coloneqq \mathbb{E}_{x \sim P}\!\left[ \log \frac{p(x)}{q(x)} \right].
\end{equation}
The CIMD quantities we define below apply to both discrete and continuous formulations of KL divergence, while our closed-form results (Section~\ref{sec: empirica}) instantiate the continuous, Gaussian setting.
One interpretation of $D(P || Q)$ is the (average) ``surprise'' when assuming $Q$ and seeing $P$.

The KL divergence between a joint probability distribution on two random variables, $P(x,y)$, and their product, $P_{X \independent Y} = P(x)P(y)$, gives ``mutual information'':
\begin{equation} \label{eq: mi}
    I(X:Y) \coloneqq D(P || P_{X \independent Y}).
\end{equation}
Conditional mutual information uses conditional probabilities in Eq.~\eqref{eq: mi}. For additional background, see \citet{cover1999elements}.

\section{ILLUSTRATING META-DEPENDENCE} \label{sec: CI Dependence Examples}


\vspace{0em}In this section, we illustrate the geometric intuition between CI dependencies by perturbing the relationships between different variables.
In particular, we consider a joint distribution $P$ that is generated from a linear additive Gaussian noise model. By describing the joint Gaussian distributions using an arbitrary set of structural equations, 
local perturbations on equation parameters can mimic the estimation errors of empirical distributions on finite samples.
These perturbations alter the partial correlation between variables, which can result in weaker or stronger dependencies.

Intuitively, for each tuple $A,B,\bvec{C}$, the set of joint distributions $\mathcal{P}_{A \independent B \given \bvec{C}}$ that satisfy $\rho_{A,B\mid \bvec{C}} = 0$ constitutes a manifold in the space of possible joint distributions. Weaker conditional dependence then corresponds, informally, to the distribution being ``closer'' to this manifold.\footnote{This proximity is not necessarily Euclidean distance in a chosen parameterization. Section~\ref{sec: definition} will sharpen this using an information-geometric notion of closeness, which is captured by the covariance matrix for Gaussians.} A perturbation that alters the joint distribution might move it towards two such manifolds simultaneously, constituting a positive dependence, or towards one while moving away from the other, a negative dependence.
By observing how proximities to different CI manifolds change after these parameter perturbations, we obtain insights into the dependencies between different CI tests. In particular, we will demonstrate how identical causal structures with different parameters can exhibit different dependencies between their CI properties.

\subsection{Parameter Perturbations}

In linear additive Gaussian models, each variable $V$ is associated with an additive, independent Gaussian noise term $N_V$. To simplify our examples, we will assume that $N_V$ is chosen so that all random variables are standardized, meaning that $\rho_{A, B \given \bvec{C}} = \Cov(A, B \given \bvec{C})$.

Consider the following system with $\alpha\neq 0$:
\begin{equation*}
\begin{aligned}
    A &= N_A, & B &= \alpha A + N_B.
\end{aligned}
\end{equation*}
According to this characterization, $\rho_{A,B} =\alpha$. Perturbing $\alpha$ allows us to directly change the covariance between $A,B$.


We now consider two CI properties jointly and give examples of both ``positive dependence'' and ``negative dependence''. 
A \emph{positive} dependence involves a perturbation \emph{towards} one CI manifold that also moves the joint distribution \emph{towards} another CI manifold. 
In contrast, a \emph{negative} dependence involves a perturbation \emph{towards} one CI manifold that moves the joint distribution \emph{away from} another.

\subsection{Positive Dependence}\label{sec:pos-example}

\vspace{0em}Consider the following structural equations that capture a fully connected DAG: $A \rightarrow B \rightarrow C, A \rightarrow C$,
\begin{equation}\label{eq: neg dependence}
    \begin{aligned}
        A &= N_A, & B &= \alpha_1 A + N_B, & C &= \alpha_2 A + \beta B + N_C.
    \end{aligned}
\end{equation}
The variances of the noise terms are again chosen so that the random variables are standardized. Writing each variable in terms of $N_A, N_B, N_C$,
\begin{equation*}
    \begin{aligned}
        A &= N_A,\\
        B &= \alpha_1 N_A + N_B,\\
        C &= \alpha_2 N_A + \beta \alpha_1 N_A + \beta N_B + N_C.
    \end{aligned}
\end{equation*}
Using $\Var(N_B) = 1 - \alpha_1^2$, the three correlations for this model are,
\begin{equation*}
    \begin{aligned}
        \rho_{A,B} &= \alpha_1, &
        \rho_{A,C} &= \alpha_2 + \alpha_1 \beta, &
        \rho_{B,C} &= \alpha_1\alpha_2 + \beta.
    \end{aligned}
\end{equation*}

For the first example, let us assume $\alpha_2 = 0$, which corresponds to a simplified DAG $\cG: A \rightarrow B \rightarrow C$. Notice that a perturbation lowering $\alpha_1$ consequently lowers the dependencies between $A,B$ and $A,C$ simultaneously by reducing the magnitudes of both $\abs{\rho_{A,B}}$ and $\abs{\rho_{A,C}}$. This constitutes a \emph{positive dependence}.



\subsection{Negative Dependence} \label{sec: neg dependence}

\begin{figure*}
    \centering
    \begin{tikzpicture}
        \node (diagram) at (0, 0) {\scalebox{.55}{\begin{tikzpicture}
\begin{axis}[ 
    axis lines = middle, 
    xlabel = {$\beta$}, 
    ylabel = {$\alpha_2$}, 
    xmin = -.5, xmax = .5, 
    ymin = -.5, ymax = .5,
]
    \addplot[red, thick, domain=-.5:.5, name path=A] {-.5*x};
    \addplot[blue, thick, domain=-.5:.5, name path=B] {-2*x};
    \addplot [opacity=.4, color=gray] fill between [
        of=A and B,soft clip={domain=-.5:.5},
    ];
    \node[red] at (axis cs:-0.4, .1) {$\rho_{A,C} = 0$};
    \node[blue] at (axis cs:0.11, -0.45) {$\rho_{B, C} = 0$};
    \node[text width=2cm, align=center] (nd) at (axis cs:.33, -.3) {Negative Dependence};
    \node[text width=2cm, align=center] (nd) at (axis cs:-.33, .3) {Negative Dependence};
    \node[text width=2cm, align=center] (pd) at (axis cs:0.3, .3) {Positive Dependence};
    \node[text width=2cm, align=center] (pd2) at (axis cs:-0.3, -.3) {Positive Dependence};
\end{axis}
\end{tikzpicture}}};
    \node (fscid) at (11, 0) {\includegraphics[width=.33 \textwidth]{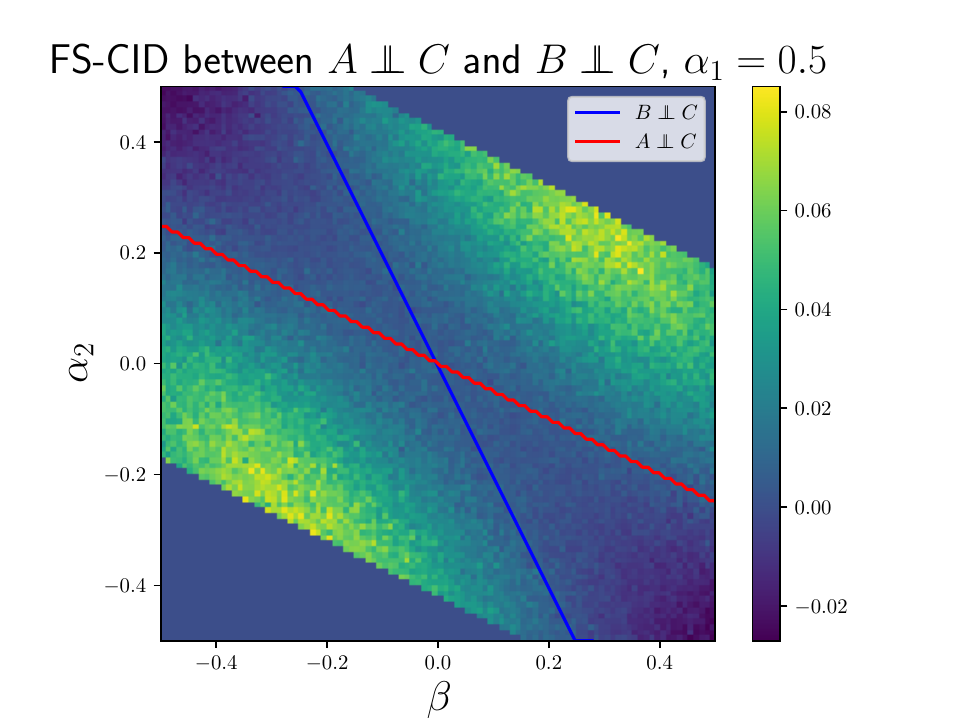}};
    \node (CIMD) at (5.5, 0) {\includegraphics[width=.32 \textwidth]{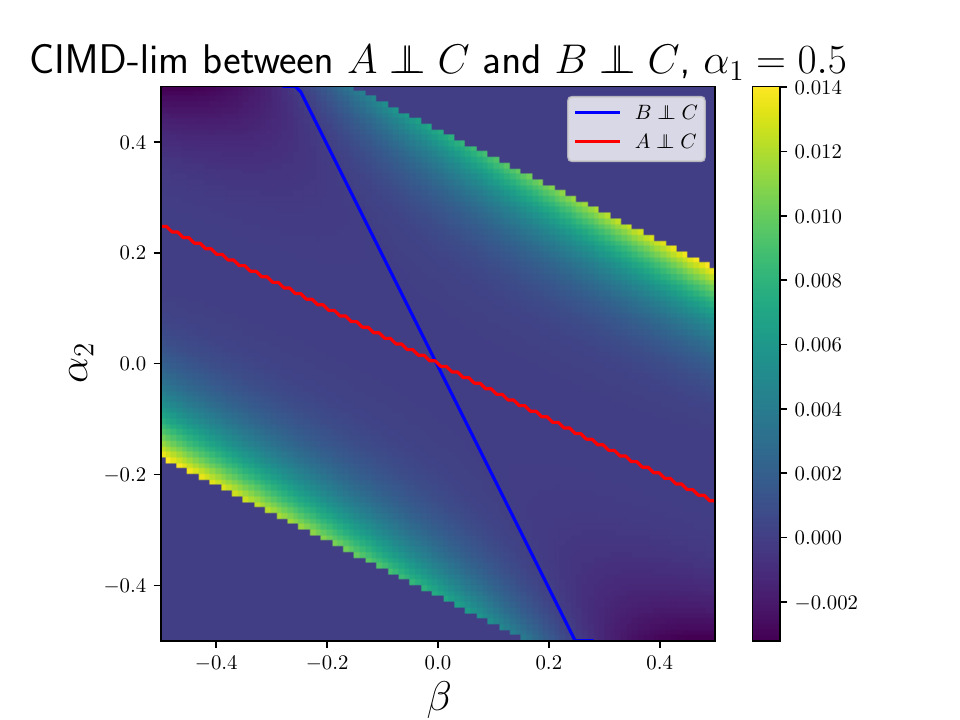}};
    \node at (0, -2.5) {(a)};
    \node at (5.5, -2.5) {(b)};
    \node at (11, -2.5) {(c)};
    \end{tikzpicture}
    \caption{(a) Dependence of CI properties for structural equations given in Eq.~\eqref{eq: neg dependence} when $\alpha_1 = .5$. The shaded region between the red and blue lines corresponds to models where moving towards blue moves \emph{away} from red. Hence, the shaded region has a ``negative'' dependence between $A \independent C$ and $B \independent C$, while the unshaded region has a ``positive'' dependence.
    (b) The introduced metric, CIMD (Section~\ref{sec: definition}), measuring the meta-dependence between CI-tests, computed from the covariance matrix. (c) Estimation of CIMD with finite samples (Sections~\ref{sec: empirica}~and~\ref{sec: empirical}).}
    \label{fig:pos and neg corrleation}
\end{figure*}

Now, consider the joint distribution $P$ generated by 
\begin{equation}\label{eq:neg-example}
    \begin{aligned}
        \alpha_1 &= 0.5, & \alpha_2 &= 0.3, & \beta &= -0.3.
    \end{aligned}
\end{equation}
The value $\alpha_2 = 0.3$ fixes a specific base distribution $P$; treating $\alpha_2$ as an indeterminate in a neighborhood of this value lets us analyze how small perturbations move $P$ relative to nearby CI manifolds. Concretely, we track the change in $\rho_{B,C}$ and $\rho_{A,C}$ as $\alpha_2$ is perturbed around $0.3$:
\begin{equation}
    \begin{aligned}
        \rho_{A,C} &= \alpha_2 -0.15 = 0.15,\\
        \rho_{B,C} &= 0.5\alpha_2 - 0.3 = -0.15.
    \end{aligned}
\end{equation}
Notice that $\rho_{A,C} > 0$ while $\rho_{B,C} < 0$. As a result, changing $\alpha_2$ moves the absolute values of the two correlations in opposite directions. Therefore, in the neighborhood of $P$ as given by the parameters in Equation~\eqref{eq:neg-example}, perturbations of $\alpha_2$ that shift $P$ towards $A\independent C$ are unlikely to move the distribution ``closer'' to $B\independent C$.



\paragraph{Remarks.} Our example of positive dependence in Section~\ref{sec:pos-example} holds for \emph{all} parameter specifications with $\alpha_2 = 0$ --- i.e., distributions that follow the DAG structure $A \rightarrow B \rightarrow C$. That is, $A\independent B$ and $A\independent C$ co-occur when perturbing $\alpha_1$ regardless of the specification of $\beta >0$.
In contrast, the example of negative dependence is specific to perturbations of the distribution $P$ specified by Eq.~\eqref{eq:neg-example}. For example, consider an alternative joint distribution $Q$ generated using
\begin{equation}\label{eq:pos-example}
    \begin{aligned}
        \alpha_1 &= 0.5, & \alpha_2 &= 0.3, & \beta &= 0.3.
    \end{aligned}
\end{equation}
Now, both correlations are positive, meaning that perturbations in $\alpha_2$ move their absolute values in the same direction. Hence, although the same graphical model underlies $P$ and $Q$, these distributions differ in their dependencies between (local) conditional independencies. More generally, Figure~\ref{fig:pos and neg corrleation} (a) illustrates the space of possible models generated by different specifications of Eq.~\eqref{eq: neg dependence}, with regions of positive and negative dependence annotated. We visualize the manifolds of independencies using a red line for $A\independent C$ and a blue line for $B \independent C$.

It is clear that we cannot rely solely on the graph structure when studying the dependence between CI tests under a random perturbation. Instead, we need to quantify how measures of dependence change relative to local perturbations of the joint distribution. 
In the next section, we will define an information-projection-based quantity to formally measure shared information between two CI tests. This quantity will not require access to the parameters of the structural equations.


\section{QUANTIFYING META-DEPENDENCE} \label{sec: definition}
We now make meta-dependence precise. CIMD quantifies the shared information between two conditional independence tests via moment projections and requires no access to the causal graph. Section~\ref{sec: empirical} will then CIMD to the co-occurrence of test outcomes under finite-sample uncertainty.

\subsection{Projecting into Independence}

To define CIMD, we first relate a joint distribution to a single conditional independence through moment projections.

Consider a graphical model on the two vertices $\{X, Y\}$. Restricting attention to directed acyclic graphs without latent confounders, a joint distribution $P$ with $X \not\independent Y$ is modeled as $X \rightarrow Y$ or $X \leftarrow Y$, while a distribution with $X \independent Y$ is modeled by the \emph{empty graph} on $\{X, Y\}$ (no edges). The M-projection of $P$ onto this empty graph is the product of its marginals, $\Pi_{X\independent Y}(P)$ (defined below), and the mutual information measures how far $P$ sits from it in KL divergence:
\begin{equation*}
    I(X:Y) = D(P || \Pi_{X \independent Y}(P)).
\end{equation*}

This product-of-marginals construction is special to \emph{marginal} independence; for a conditional independence $A \independent B \given \bvec{C}$ we instead obtain the projection from a general characterization of projections onto graphs. The M-projection\footnote{See \citet{koller2009probabilistic} Ch 8.5.} of $P$ onto a set of distributions $\mathcal{P}$ is its KL-closest member,
\begin{equation}\label{def: independence projection}
    \Pi_{\mathcal{P}}(P) \coloneqq \argmin_{Q \in \mathcal{P}} D(P || Q),
\end{equation}
and when $\mathcal{P} = \mathcal{P}_{\mathcal{G}}$ is the set of distributions Markov to a graph $\mathcal{G}$, it takes a factorized form.

\begin{theorem}[Thm 8.7 in \citet{koller2009probabilistic}]
\label{thm: koller project}
    A moment projection of $P$ onto $\G$ is given by the factorization implied by $\G$ using the conditional probabilities from $P$:
    \begin{equation*}
        \big[\Pi_{\G}(P)\big](\bvec{v}) = \prod_{V_i \in \bvec{V}} P(v_i \given \pa(v_i)).
    \end{equation*}
\end{theorem}
The left-hand side $\Pi_{\G}(P)$ is the projected distribution (at assignment $\bvec{v}$) and the right-hand side is its factorization using the conditionals of the \emph{original} $P$.

Theorem~\ref{thm: koller project} projects onto a \emph{graph}, not a single conditional independence. To reuse it for a CI, we introduce an \emph{auxiliary graph}: a placeholder DAG chosen only to induce the target independence. For $A \independent B \given \bvec{C}$, the fork $A \leftarrow \bvec{C} \rightarrow B$ suffices.\footnote{Any Markov-equivalent choice, such as the chain $A \rightarrow \bvec{C} \rightarrow B$, gives the same projection.}

\begin{corollary}
\label{thm: CI project}
    Let $\mathcal{P}_{A \independent B \given \bvec{C}}$ be the set of distributions satisfying $A \independent B \given \bvec{C}$, and realize the projection onto it through the auxiliary graph $A \leftarrow \bvec{C} \rightarrow B$. With $\bvec{X} = \bvec{V} \setminus (\{A, B\} \cup \bvec{C})$ the remaining variables, the marginal of $\Pi_{A \independent B \given \bvec{C}}(P)$ over $A, B, \bvec{X}$ (with $\bvec{C}$ integrated out) is
    \begin{align*}
        \big[\Pi_{A \independent B \given \bvec{C}}(P)\big]&(a, b, \bvec{x}) =\\ &\E_{\bvec{C}}[P(a \given \bvec{C}) P(b \given \bvec{C}) P(\bvec{x} \given a, b, \bvec{C})].
    \end{align*}
\end{corollary}
\begin{proof}
    $A \independent B \given \bvec{C}$ implies the factorization $P(\bvec{V}) = P(\bvec{C}) P(A \given \bvec{C}) P(B \given \bvec{C}) P(\bvec{X} \given A, B, \bvec{C})$, i.e., the auxiliary graph $A \leftarrow \bvec{C} \rightarrow B$, to which we apply Theorem~\ref{thm: koller project}.
\end{proof}
Crucially, this projection requires no knowledge of the underlying graphical structure.

\subsection{CI Meta-Dependence}
To measure the shared information between two CI tests, we ask whether the conditional mutual information for one test changes after projecting onto the other.
\begin{definition}[CIMD]
Consider two conditional independencies:
\begin{align*}
(T_1) \hspace{.5em} A &\independent B \given \bvec{C},   &   (T_2) \hspace{.5em} A' &\independent B' \given \bvec{C'}.
\end{align*}
The \emph{CI Meta-Dependence (CIMD)} between them on an empirical distribution $P$ is
\begin{equation*}
    \CIMD(T_1, T_2, P) \coloneqq I_{P}(A : B \given \bvec{C}) - I_{\Pi_{T_2}(P)}(A : B \given \bvec{C}).
\end{equation*}
Here $I_P$ is the conditional mutual information under $P$, and $I_{\Pi_{T_2}(P)}$ is the same functional under $\Pi_{T_2}(P)$, the projection of $P$ onto the null of $T_2$ (Eq.~\eqref{def: independence projection}). The two tests play different roles: $T_1$ fixes \emph{which} mutual information is measured, $I(A : B \given \bvec{C})$, while $T_2$ fixes the projection applied beforehand. The variables need not be disjoint --- $A, B, \bvec{C}$ and $A', B', \bvec{C}'$ may overlap, and the primed variables are not separate copies of the unprimed ones.
\end{definition}

Reading a moment projection as the minimal perturbation of $P$ onto a CI property, CIMD asks whether that perturbation also moves toward a second CI property. Zero CIMD means the two tests are ``orthogonal'': the CI manifolds are perpendicular, so moving onto one leaves the distance to the other unchanged. Positive CIMD means projecting onto $T_2$ also moves $P$ toward $T_1$'s null, so the manifolds point the same way; negative CIMD, which is rarer (Section~\ref{sec: empirical}), means they lie on opposite sides of $P$. Figure~\ref{fig:pos and neg corrleation} illustrates both cases.

\paragraph{Remarks.} We note three properties of CIMD. First, it is asymmetric: $\CIMD(T_1, T_2, P) \neq \CIMD(T_2, T_1, P)$ in general. In the DAG $A \rightarrow B \rightarrow C \rightarrow D$, perturbing toward $B \independent C$ forces $A \independent D$, since severing $B$--$C$ disconnects $A$ from $D$; but perturbing toward $A \independent D$ need not force $B \independent C$, as it may instead arise from a $C \independent D$ faithfulness violation. This fixes the order of computation: to ask whether $T_1$ is \emph{redundant} once $T_2$ holds --- e.g., when $T_2$'s outcome was already spent earlier in a sequence of tests --- one computes $\CIMD(T_1, T_2, P)$, where a large positive value means $T_1$ adds little beyond $T_2$.

Second, CIMD is related to, but distinct from, Euclidean distance in the parameter space of the structural equation model: in Figure~\ref{fig:pos and neg corrleation}, the moment projections do not land on the nearest point of the red and blue lines but reach them by zeroing out correlations.

Third, CIMD depends on $P$: the same pair of tests can yield positive CIMD for one distribution and negative for another (Section~\ref{sec: neg dependence}).

We use KL divergence because its M-projection has a closed form in exponential families (Section~\ref{sec: empirica}) and it carries the directional ``surprise'' reading of Section~\ref{sec:prelim}, matching our intended meaning --- how far a distribution must move to satisfy a test. A symmetric divergence could be substituted, at the cost of this tractability.

\subsection{A Gaussian Closed Form}\label{sec: empirica}
For multivariate Gaussians, both the moment projection and CIMD can be computed in closed form directly from the covariance matrix, with no access to the raw data. Let $\Sigma$ be the covariance of $P$ over $\bvec{A}, \bvec{B}, \bvec{C}, \bvec{X}$ (with $\bvec{X}$ the remaining variables), and let $\Sigma^{\independent}$ be the covariance of the projection $\Pi_{\bvec{A} \independent \bvec{B} \given \bvec{C}}(P)$. By Corollary~\ref{thm: CI project}, $\Sigma^{\independent}$ agrees with $\Sigma$ except in the two blocks involving $\bvec{A} \cup \bvec{B}$ and $\bvec{X}$: one zeroes the $\bvec{A}, \bvec{B}$ conditional covariance given $\bvec{C}$, the other propagates that change through $P(\bvec{X} \given \bvec{A}, \bvec{B}, \bvec{C})$ (full construction in Appendix~\ref{apx: gaussian derivation}). Letting $\Phi$ and $\Phi^{\independent}$ denote the conditional covariance of $\bvec{Z} = \bvec{A} \cup \bvec{B}$ given $\bvec{C}$ under $P$ and under $\Pi_{\bvec{A} \independent \bvec{B} \given \bvec{C}}(P)$ respectively, the Gaussian mutual-information formula gives
\begin{equation}\label{eq: gaussian cimd}
\begin{aligned}
    \CIMD(T_1, T_2) = &\frac{1}{2}\log{\left(\frac{\det(\Phi_{\bvec{A},\bvec{A}})\det(\Phi_{\bvec{B},\bvec{B}})}{\det(\Phi)}\right)} \\
    - &\frac{1}{2}\log{\left(\frac{\det(\Phi^{\independent}_{\bvec{A},\bvec{A}})\det(\Phi^{\independent}_{\bvec{B},\bvec{B}})}{\det(\Phi^{\independent})}\right)},
\end{aligned}
\end{equation}
where $\Phi_{\bvec{A},\bvec{A}}$ and $\Phi_{\bvec{B},\bvec{B}}$ reduce to scalars when $A$ and $B$ are single variables. Both terms depend on $P$ only through $\Sigma$, so CIMD is a function of the covariance matrix alone.

\section{EMPIRICAL TESTS}\label{sec: empirical}

\vspace{0em}CIMD does not directly address finite sample deviations. However, KL divergence still measures statistical difference: a small $D(P || Q)$ indicates that samples drawn from $P$ are, on average, not very surprising under $Q$ --- equivalently, $Q$ assigns high likelihood to data typical of $P$. In keeping with the asymmetry of KL, this is a directional statement and does not imply the reverse. Motivated by this intuition, we give experiments on both synthetic and real-world data that show CIMD aligns with the dependence between outcomes of CI-tests under finite-data uncertainty. \footnote{Code can be found on \href{https://github.com/honeybijan/CIMD_experiments}{GitHub}.}

\begin{figure*}[!h]
\centering
(a)\includegraphics[width=.44\textwidth]{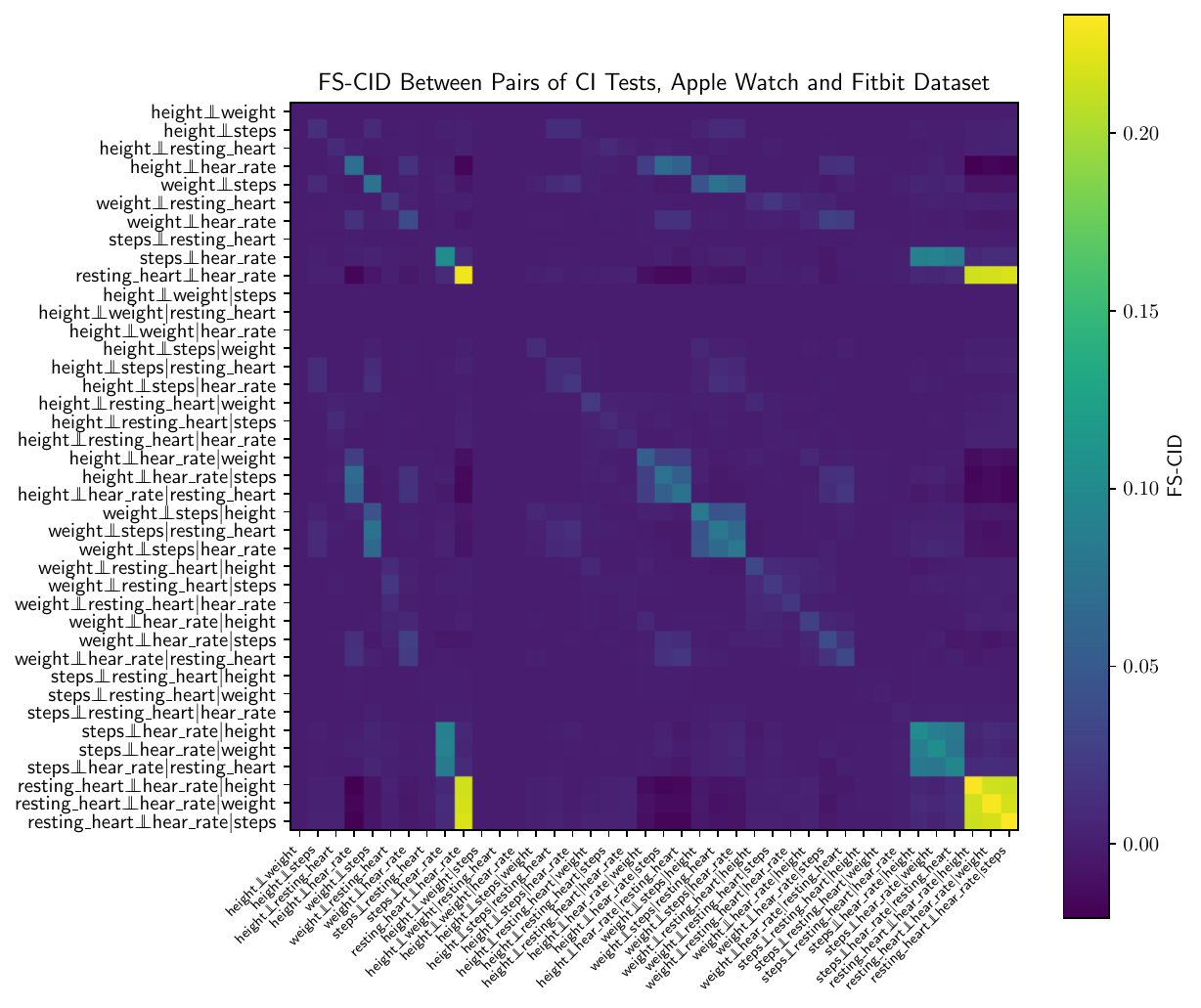}
(b)\includegraphics[width=.44\textwidth]{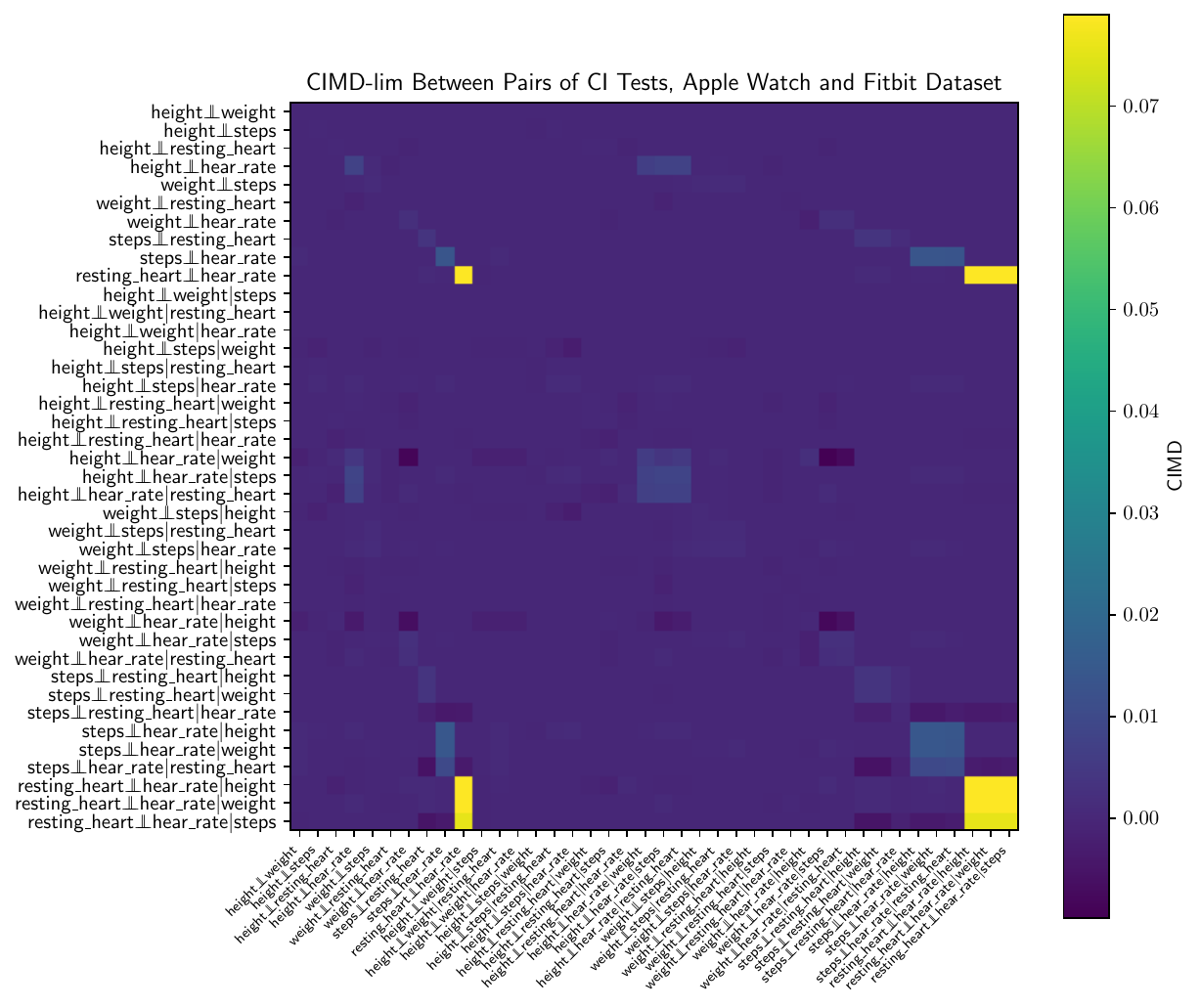}
\caption{Meta-dependencies of CI-tests as measured by FS-CID in (a) and CIMD-lim in (b).} \label{fig: apple watch}
\end{figure*}

\paragraph{FS-CID.}
We will use ``Finite-Sample CI Dependence'' (FS-CID) to measure the dependence between two CI-tests due to finite-sample uncertainty. FS-CID bootstraps perturbations of the true distribution to empirical ones and keeps track of the co-occurrence of CI-test significance. To do this, we randomly generate $1000$ smaller datasets. For synthetic data, these smaller datasets come from sampling a multivariate Gaussian variable. For real-world data, these datasets are random sub-samples from a larger dataset.

\vspace{0em}For each of these datasets, we compute the result of a CI-test using partial correlation with a Fisher-Z transformation.\footnote{This hypothesis test is implemented in \texttt{conditional-independence} \citep{squires2018causaldag}. All experiments are conducted on 1 CPU.} If $t_1$ corresponds to a failure to reject test 1 and $t_2$ failure to reject test 2, then we report the FS-CID as
\begin{equation} \label{eq: fscid}
    \text{FS-CID} \coloneqq \hat{P}(t_1, t_2) - \hat{P}(t_1)\hat{P}(t_2),
\end{equation}
where $\hat{P}$ are empirical estimates from $1000$ datasets.

\paragraph{Limited CIMD.}
When either null hypothesis is very likely to be rejected (i.e., strong dependence), we will likely have $\hat{P}(t)=0$ for an FS-CID of $0$. In contrast, CIMD still projects the distribution onto that (very likely to reject) CI-test, likely giving a non-zero value. In such cases, CIMD is uninformative because the distribution is far from either CI property; the interesting meta-dependence instead occurs near the CI-test violations, where tests actually flip between rejection and non-rejection under resampling. To focus on this regime, we zero CIMD out whenever either mutual information term is large ($> 0.1$ nats), and call the result ``CIMD-lim.'' The cutoff of $0.1$ nats is a deliberately small amount of (conditional) mutual information: a term exceeding it corresponds to a dependence strong enough that the associated CI-test almost never fails to reject in FS-CID, so its meta-dependence is not resolved by the finite-sample simulation and would only add noise to the comparison.

\subsection{Synthetic Data}
For synthetic data, we calculate FS-CID by generating $1000$ datasets of size $20$ using the generative model specified by the structural equations in Eq.~\eqref{eq: neg dependence} with $\alpha_1=1$ and $\alpha_2, \beta \in [-1, 1]$.\footnote{We use relatively small (20 samples) dataset sizes in order to induce a sufficient perturbation from the true distribution to the empirical one to resolve a dependence.} These parameters correspond exactly to those used in Figure~\ref{fig:pos and neg corrleation} (a), with results shown in (b) and (c). 

\vspace{0em}Unlike the single-parameter perturbations studied in Section~\ref{sec: CI Dependence Examples}, finite-sample deviations extend in any direction, possibly affecting more than one parameter. Nonetheless, the regions of positive and negative dependence match up for all three plots.
We provide additional synthetic experiments on different graphical structures in Appendix~\ref{apx: other graph experiments}.

\subsection{Real-World Data}
To demonstrate the relevance of CIMD in practical studies, we run experiments on the ``California Housing'' \citep{pace1997sparse}, ``Apple Watch and Fitbit'' \citep{DVN/ZS2Z2J_2020}, and ``Auto MPG'' \citep{auto_mpg_9} datasets. These real-world datasets were chosen arbitrarily and have few continuous covariates, keeping our experiment size within reason.
FS-CID is computed in these settings by sub-sampling $50$ data points from these datasets $1000$ times.\footnote{We use a relatively small sub-sample size (50 points) in order to induce a sufficient perturbation from the true distribution to the empirical one to resolve a dependence.}

\vspace{0em}We give the results for the ``Apple Watch and Fitbit'' dataset in Figure~\ref{fig: apple watch}. Notice that CIMD-lim and FS-CID recover very similar meta-dependencies across the many conditional-independence tests. This dataset has relatively sparse dependence, giving an easy-to-recognize pattern. Much of the recovered meta-dependence concentrates on pairs of CI-tests that share variables; this is expected, since shared variables are precisely where a moment projection most strongly perturbs the dependencies that both tests probe. We therefore read Figure~\ref{fig: apple watch} as evidence that CIMD tracks FS-CID wherever meta-dependence is present, rather than as a claim about how prevalent such dependence is across arbitrary pairs of tests.
The results for the other two datasets are given in Appendix~\ref{apx: other real world}. These two datasets have more complicated meta-dependencies but still share signal across the two metrics.

\subsection{Tuning Causal Discovery}\label{sec: causal discovery}
As a proof of concept for using CIMD inside an algorithm, we apply it to the multiple-testing correction in constraint-based causal discovery. The PC algorithm removes an edge $\{A,B\}$ whenever \emph{some} conditioning set $\bvec{S}$ yields $A \independent B \given \bvec{S}$, so each candidate edge is associated with an entire family of CI tests. A Bonferroni correction controls the family-wise error rate by dividing the significance level $\alpha$ by the total number of such tests, implicitly treating them as independent. Many of these tests are highly redundant, however: for a fixed edge, $A \independent B \given \bvec{S}$ and $A \independent B \given \bvec{S}'$ probe overlapping dependencies. We use CIMD to estimate this redundancy and relax the correction accordingly.

For each edge, let $n_{\mathrm{edge}}$ be the number of conditioning sets it is tested against. We estimate a mean redundancy $\bar{r} \in [0,1]$ by averaging, over sampled pairs of conditioning sets, the CIMD of Section~\ref{sec: definition} normalized by $I_P(a\,{:}\,b \given \bvec{S})$ and gated to the small-dependence regime exactly as the CIMD-lim of Section~\ref{sec: empirical}. We then replace $n_{\mathrm{edge}}$ with an \emph{effective} test count $n_{\mathrm{eff}} = n_{\mathrm{edge}}\,(1-\bar{r})$ in the Bonferroni denominator, so that edges with redundant conditioning-set tests receive a less stringent threshold; when $\bar{r}=0$ the rule reduces exactly to Bonferroni. Because the normalized CIMD is a function of the covariance matrix (Section~\ref{sec: empirica}), the entire correction is computed from summary statistics of the data.

\begin{figure*}[t]
\centering
\includegraphics[width=\textwidth]{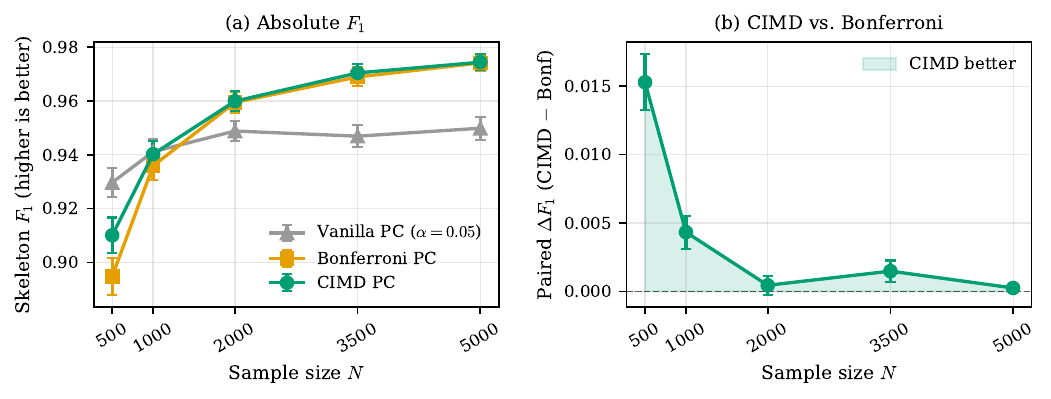}
\caption{Using CIMD to relax the multiple-testing correction in PC-based causal discovery, on sparse linear-Gaussian DAGs ($p=18$, averaged over $100$ random graphs; error bars are $\pm 1$ SEM). (a) Skeleton $F_1$ for uncorrected PC, Bonferroni-corrected PC, and CIMD-corrected PC; both corrections overtake uncorrected PC as $N$ grows, while uncorrected PC's precision plateaus. (b) Paired difference $\Delta F_1 = F_1^{\mathrm{CIMD}} - F_1^{\mathrm{Bonf}}$ per graph: CIMD-PC improves on Bonferroni-PC at small $N$ and coincides with it for $N \geq 2000$.}
\label{fig: cimd pc}
\end{figure*}

We evaluate on sparse linear-Gaussian DAGs ($p=18$ variables, $\approx 1$ expected parent per node, conditioning sets up to size $3$), estimating the skeleton with a partial-correlation (Fisher-Z) test at $\alpha = 0.05$ and averaging over $100$ random graphs at each sample size. We compare three edge-thresholding rules applied to the \emph{same} skeleton search: uncorrected (``vanilla''), Bonferroni, and the CIMD-relaxed correction. Figure~\ref{fig: cimd pc} reports the skeleton $F_1$. The redundancy CIMD detects is substantial ($\bar{r} \approx 0.6$--$0.7$), so it treats only $30$--$40\%$ of each edge's conditioning-set tests as effective and relaxes the per-edge threshold accordingly. Two patterns emerge. First, uncorrected PC trades precision for recall: a fixed $\alpha$ leaves a roughly constant false-positive rate that more data does not remove, so its precision stays near $0.93$ at every sample size while its recall climbs; its $F_1$ therefore rises only modestly (from $0.93$ to $0.95$) and plateaus below both corrections, which reach precision $\approx 0.99$ and $F_1 \approx 0.97$ by $N=5000$. The corrections overtake uncorrected PC once $N \geq 2000$; at the smallest samples their conservatism costs enough recall that uncorrected PC retains the highest $F_1$. Second, and consistent with the modest claim of Section~\ref{sec:concl}, CIMD-PC never underperforms Bonferroni-PC and improves on it exactly where the two differ most --- the small-sample regime: at $N=500$ it recovers more true edges at matched precision, giving $\Delta F_1 = +0.015 \pm 0.002$ and $0.47 \pm 0.06$ fewer skeleton errors, with the gap closing to zero by $N \geq 2000$. This setting is deliberately favorable to correction --- a sparse graph with a large, redundant family of conditioning-set tests --- so we present it as evidence that CIMD's redundancy estimate can be turned into a usable, distribution-aware multiplicity correction, not as a claim of universal improvement.

\section{CONCLUSION}\label{sec:concl}
This paper introduces the study of CI Meta-Dependence (CIMD), an information-theoretic measure of the shared information between two tests for conditional independence. Our concrete contributions are twofold. First, we show empirically --- on both synthetic and real-world data (Section~\ref{sec: empirical}) --- that CIMD aligns with FS-CID, the co-occurrence of CI-test outcomes under finite-sample uncertainty. Second, we derive a closed form for CIMD in the multivariate Gaussian case (Section~\ref{sec: empirica}) that is computable directly from a covariance matrix, making the measure tractable in practice. The definition itself (Section~\ref{sec: definition}) is general and applies to any distribution that factorizes over a graph; only the closed-form computation is specific to the Gaussian setting.

Beyond these results, CIMD suggests several directions for algorithms that rely on sequences of CI tests. We describe them below, distinguishing what the present paper supports from what remains open.

\paragraph{Significance Thresholds.}
Constraint-based causal discovery employs many CI tests, many of which correspond to removing an edge between two variables. Multiple-testing corrections (e.g., Bonferroni) typically treat these tests as if they were independent, tightening each significance threshold in proportion to the total number of tests. When two tests share information --- and therefore tend to co-occur --- treating them as independent is overly conservative. CIMD quantifies this redundancy directly from the distribution, which suggests replacing the uniform correction with a redundancy-aware one. We make a deliberately modest claim: \emph{when a multiple-testing correction is going to be applied anyway}, accounting for the redundancy measured by CIMD is a strict improvement over assuming independence between tests. We demonstrate this concretely in Section~\ref{sec: causal discovery}: replacing Bonferroni's test count with a CIMD-based effective count matches or slightly improves the skeleton recovered by the PC algorithm, at no cost in precision. We do not claim it delivers exact finite-sample error control, which is infeasible even for a single CI test \citep{shah_hardness_2020}.

\paragraph{Validating Causal Discovery.}
Validating causal discovery requires a distance between the true ($\mathcal{G}$) and recovered ($\hat{\mathcal{G}}$) graphs, such as the Structural Hamming Distance \citep{de2009comparison} or intervention-based metrics \citep{peters2015structural, henckel2024adjustment, wahl2025separationbased}. Because CI meta-dependence is distribution-specific, a fixed structural difference can be more or less detectable across different distributions, so such graphical error measures should be read with the distribution in mind. CIMD instead quantifies \emph{co-occurring} edge errors: a single distribution-driven mistake may matter as much as many co-occurring ones. Projecting onto a manifold for a \emph{full} graphical structure (Section~\ref{sec: definition}) could then yield distribution-aware DAG losses that focus validation on statistically significant errors.

\paragraph{Uncertainty in Causal Discovery.}
CIMD is a step toward quantifying higher-order uncertainty in causal structure learning. While methods do exist to quantify uncertainty in CI testing \citep{de2014bayesian}, it remains unclear how to propagate this uncertainty into a causal structure without subsampling as in \citet{liu2010stability}. CIMD provides a moment-based (or covariance-based) metric that aligns with sub-sampling dependence (FS-CID). Since CIMD does not require access to the raw data, CIMD may be useful in large data and security-critical settings.

\paragraph{Future Work.}
Several questions remain open. \emph{(i) Non-Gaussian computation.} Our closed form is specific to multivariate Gaussians; computing CIMD more generally requires estimating moment projections and KL divergences non-parametrically, for instance via kernel-based mutual-information estimators or normalizing-flow projections onto CI manifolds. Characterizing the difficulty and accuracy of these estimators is an important next step. \emph{(ii) A formal CIMD--FS-CID link.} The agreement between CIMD and FS-CID is, at present, established only empirically. Making it rigorous appears to require controlling the local curvature of the CI manifolds around the true distribution, whose geometry has so far been characterized mainly through its singularities \citep{uhler2013geometry}. Our empirical agreement is consistent with this curvature being mild in typical neighborhoods, but we do not prove it, and we regard closing this gap as the central open question raised by this work. \emph{(iii) Gaussian finite-sample guarantees.} The Gaussian setting is the most promising place to make progress: because the partial-correlation (Fisher-Z) test is level and has known power, one could aim for a finite-sample statement expressing the joint rejection probability of two CI tests as a function of CIMD plus an error term controlled by the distance to the true distribution.

\begin{acknowledgements} 
    We thank Murat Kocaoglu, Abhinav Kumar, Beshr Islam Bouli, and the anonymous reviewers for their helpful discussions.
    B.M.~is partially supported by the Advanced Research Concepts (ARC) COMPASS program, sponsored by the Defense Advanced Research Projects Agency (DARPA) under agreement number HR001-25-3-0212. J.Z.~was partially supported  by the Eric and Wendy Schmidt Center. C.U.~was partially supported by NCCIH/NIH (1DP2AT012345), ONR (N00014-24-1-2687), DOE-ASCR (DE-SC0023187), and the Eric and Wendy Schmidt Center at the Broad Institute.
\end{acknowledgements}

\bibliography{biblio}

\newpage

\onecolumn

\title{Meta-Dependence in Conditional Independence Testing\\(Supplementary Material)}
\maketitle

\appendix

\section{Derivation of the Gaussian Closed Form}\label{apx: gaussian derivation}
Here we derive the closed form of Eq.~\eqref{eq: gaussian cimd}. For Gaussian vectors, the moment projection and CIMD are obtained by directly modifying the covariance of $P$ over $\bvec{A}, \bvec{B}, \bvec{C}, \bvec{X}$,
\begin{equation*}
    \Sigma = \begin{bmatrix}
        \Sigma_{\bvec{A} \cup \bvec{B}, \bvec{A} \cup \bvec{B}} & \Sigma_{\bvec{A} \cup \bvec{B}, \bvec{C}} & \Sigma_{\bvec{A} \cup \bvec{B}, \bvec{X}}\\
        \Sigma_{\bvec{C}, \bvec{A} \cup \bvec{B}} & \Sigma_{\bvec{C}, \bvec{C}} & \Sigma_{\bvec{C}, \bvec{X}}\\
        \Sigma_{\bvec{X}, \bvec{A} \cup \bvec{B}} & \Sigma_{\bvec{X}, \bvec{C}} & \Sigma_{\bvec{X}, \bvec{X}}
    \end{bmatrix}.
\end{equation*}
Writing $P^{\independent} \coloneqq \Pi_{\bvec{A} \independent \bvec{B} \given \bvec{C}}(P)$, Corollary~\ref{thm: CI project} gives $P^{\independent}(\bvec{C}) = P(\bvec{C})$, $P^{\independent}(\bvec{A} \given \bvec{C}) = P(\bvec{A} \given \bvec{C})$, $P^{\independent}(\bvec{B} \given \bvec{C}) = P(\bvec{B} \given \bvec{C})$, and $P^{\independent}(\bvec{X} \given \bvec{A}, \bvec{B}, \bvec{C}) = P(\bvec{X} \given \bvec{A}, \bvec{B}, \bvec{C})$. Hence the covariance $\Sigma^{\independent}$ of $P^{\independent}$ matches $\Sigma$ except in the sub-matrices $\Sigma^{\independent}_{\bvec{A} \cup \bvec{B}, \bvec{A} \cup \bvec{B}}$ and $\Sigma^{\independent}_{\bvec{X}, \bvec{X}}$.

First, $\Sigma_{\bvec{A} \cup \bvec{B}, \bvec{A} \cup \bvec{B}}$ is changed so that $\bvec{A} \independent \bvec{B} \given \bvec{C}$, by setting the conditional covariance (the Schur complement of block $\bvec{C}$) to zero and rearranging:
\begin{equation*}
    \Sigma^{\independent}_{\bvec{A} \cup \bvec{B}, \bvec{A} \cup \bvec{B}} = \Sigma_{\bvec{A} \cup \bvec{B}, \bvec{C}}(\Sigma_{\bvec{C}, \bvec{C}})^{-1}\Sigma_{\bvec{C}, \bvec{A} \cup \bvec{B}}.
\end{equation*}
Next we regenerate $\Sigma_{\bvec{X}, \bvec{X}}$ using the new distribution on $\bvec{A}, \bvec{B}$ and the same conditional $P(\bvec{X} \given \bvec{A}, \bvec{B}, \bvec{C})$. With $\overline{\bvec{X}} \coloneqq \bvec{A} \cup \bvec{B} \cup \bvec{C}$,
\begin{equation*}
    \Sigma_{\overline{\bvec{X}}, \overline{\bvec{X}}}^{\independent} = \begin{bmatrix}
        \Sigma^{\independent}_{\bvec{A} \cup \bvec{B}, \bvec{A} \cup \bvec{B}} & \Sigma_{\bvec{A} \cup \bvec{B}, \bvec{C}} \\
        \Sigma_{\bvec{C}, \bvec{A} \cup \bvec{B}} & \Sigma_{\bvec{C}, \bvec{C}}
    \end{bmatrix},
\end{equation*}
and, when invertible,
\begin{equation*}
\begin{aligned}
    \Omega_{\overline{\bvec{X}}, \overline{\bvec{X}}} &\coloneqq (\Sigma^{\independent}_{\overline{\bvec{X}}, \overline{\bvec{X}}})^{-1} - \Sigma_{\overline{\bvec{X}}, \overline{\bvec{X}}}^{-1},\\
    \Sigma^{\independent}_{\bvec{X}, \bvec{X}} &=  \Sigma_{\bvec{X}, \bvec{X}} + \Sigma_{\bvec{X}, \overline{\bvec{X}}}\Omega_{\overline{\bvec{X}}, \overline{\bvec{X}}}\Sigma_{ \overline{\bvec{X}}, \bvec{X}}.
\end{aligned}
\end{equation*}
Finally, we form the conditional covariance of $\bvec{Z} = \bvec{A} \cup \bvec{B}$ given $\bvec{C}$ for the original distribution ($\Phi$) and the projected distribution ($\Phi^{\independent}$):
\begin{equation*}
\begin{aligned}
    \Phi &= \Sigma_{\bvec{Z}, \bvec{Z}} - \Sigma_{\bvec{Z}, \bvec{C}} (\Sigma_{\bvec{C},\bvec{C}})^{-1}\Sigma_{\bvec{C},\bvec{Z}}, \\
    \Phi^{\independent} &= \Sigma^{\independent}_{\bvec{Z}, \bvec{Z}} - \Sigma^{\independent}_{\bvec{Z}, \bvec{C}} (\Sigma^{\independent}_{\bvec{C},\bvec{C}})^{-1}\Sigma^{\independent}_{\bvec{C},\bvec{Z}}.
\end{aligned}
\end{equation*}
Substituting $\Phi$ and $\Phi^{\independent}$ into the closed form for the mutual information between multivariate Gaussians yields Eq.~\eqref{eq: gaussian cimd}.

\section{Synthetic Data Experiments on Other Graph Structures}\label{apx: other graph experiments}

\begin{figure*}[!h]
\centering
(a)\includegraphics[width=.46\textwidth]{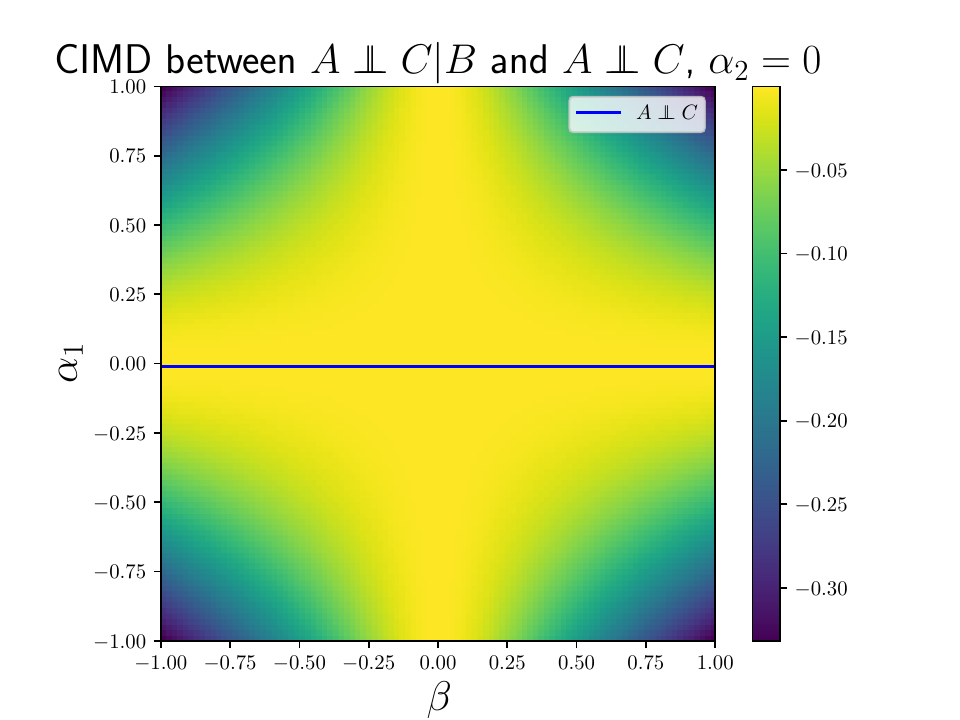}
(b)\includegraphics[width=.46\textwidth]{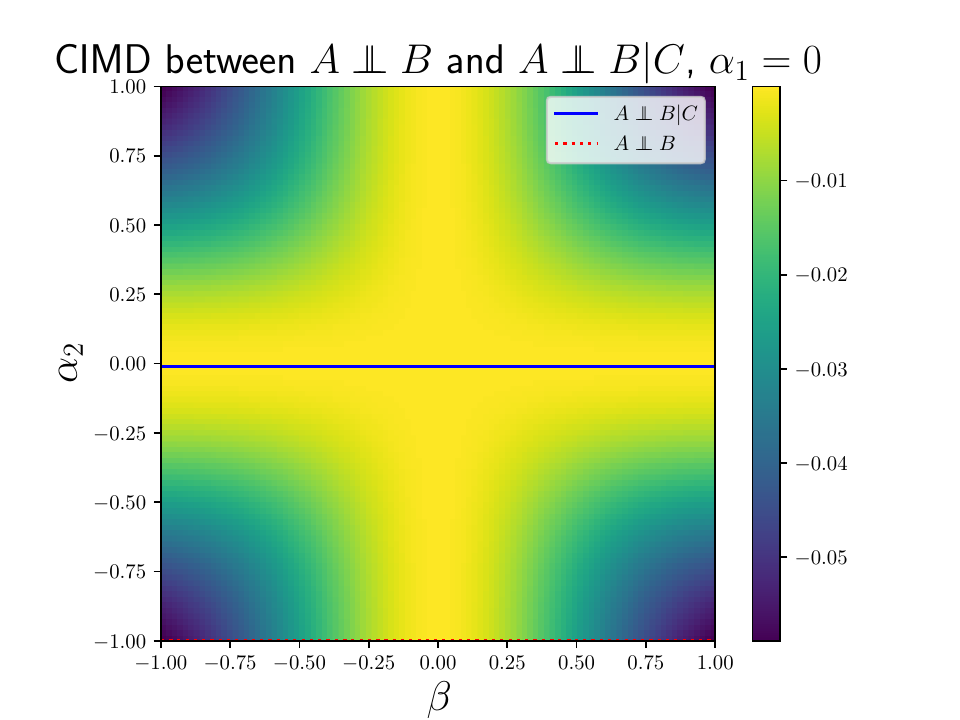}
\caption{Both (a) and (b) plot the CIMD for various parameters from the structural equations given in Eq.~\eqref{eq: neg dependence}. The effective structure in (a) is a Markov chain, and the effective structure in (b) is a collider.} \label{fig: MC and collider}
\end{figure*}

We also explore different graphical structures on three nodes by setting various parameters to $0$.

\paragraph{Markov Chain.}
When setting $\alpha_2 = 0$, we suppress the $A \rightarrow C$, giving rise to a Markov chain $A \rightarrow B \rightarrow C$, in which $A \independent C \given B$.

\paragraph{Collider.}
When setting $\alpha_1 = 0$, we suppress the $A \rightarrow B$, giving rise to a collider $A \rightarrow C \leftarrow B$, in which $A \independent B$ but $A \not \independent B \given C$.

Both the Markov chain and the collider experiments traverse surfaces in which the second test is satisfied \emph{before} any projection. For $A \rightarrow B \rightarrow C$, we start with $A \independent C \given B$ already being satisfied --- projecting onto $A \independent C$ does not take us any closer, giving a large yellow region representing a CIMD of $0$. However, once the edges between $A \rightarrow B$ and $B \rightarrow C$ get stronger (in the edges of the plot), projecting onto $A \independent C$ seems to move us away from $A \independent C \given B$, represented by the darker blue (with a negative CIMD).

Similarly, the collider begins with $A \independent B$ being satisfied, and projecting onto $A \independent B \given C$ does not affect its satisfaction (shown again by a large yellow region of $0$ CIMD). Again, stronger edges between $A \rightarrow C$ and $B \rightarrow C$ in the corners create a negative CIMD.

From these experiments, we observe that negative CIMD appears to occur between tests on the same two variables with different conditioning sets, particularly when one test holds in the ground truth distribution, and the other does not.

\section{Real-World Dataset CI Meta-Dependence}\label{apx: other real world}
\begin{figure}[!h]
    \centering

\includegraphics[width=0.49\linewidth]{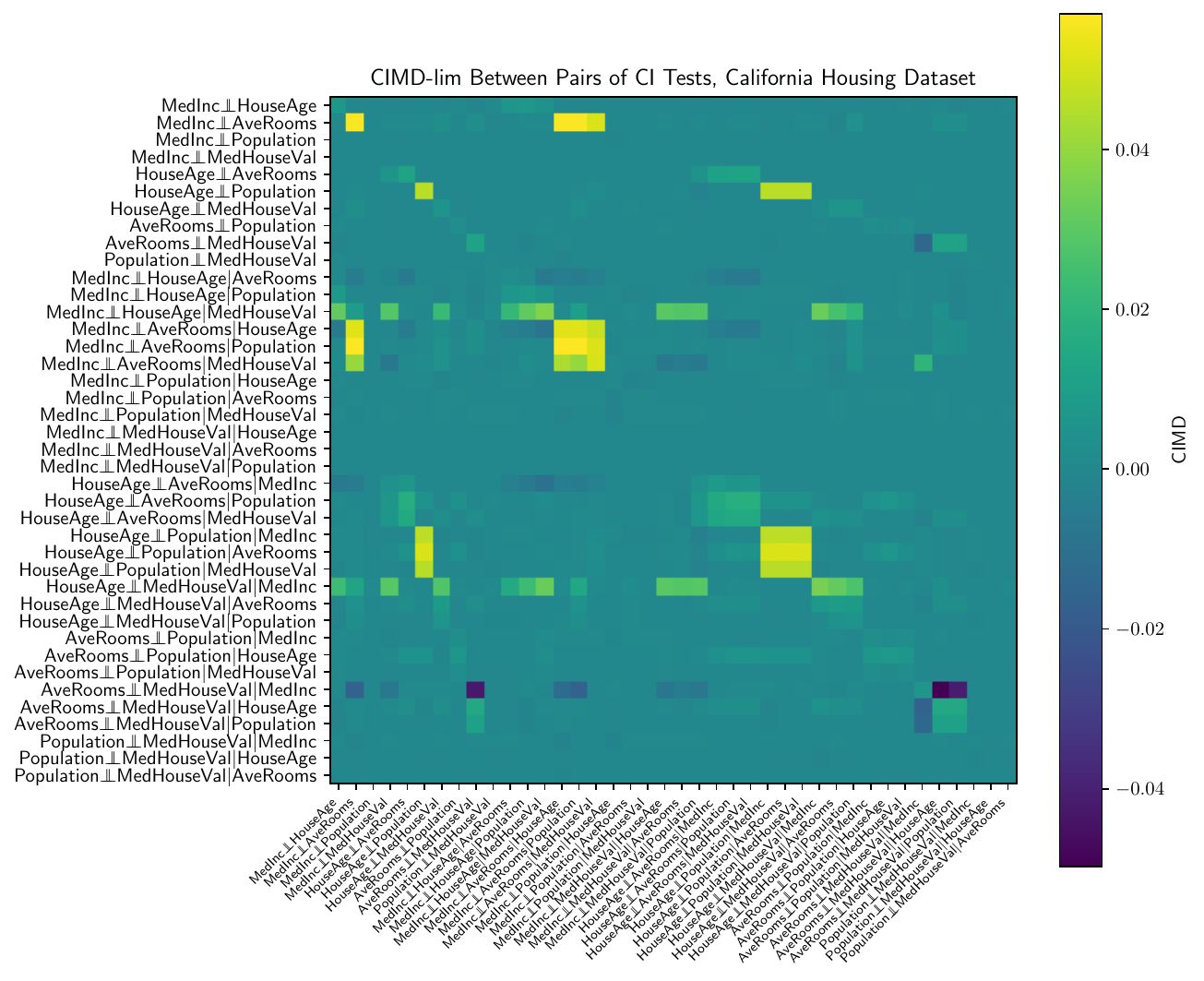}
    \includegraphics[width=0.49\linewidth]{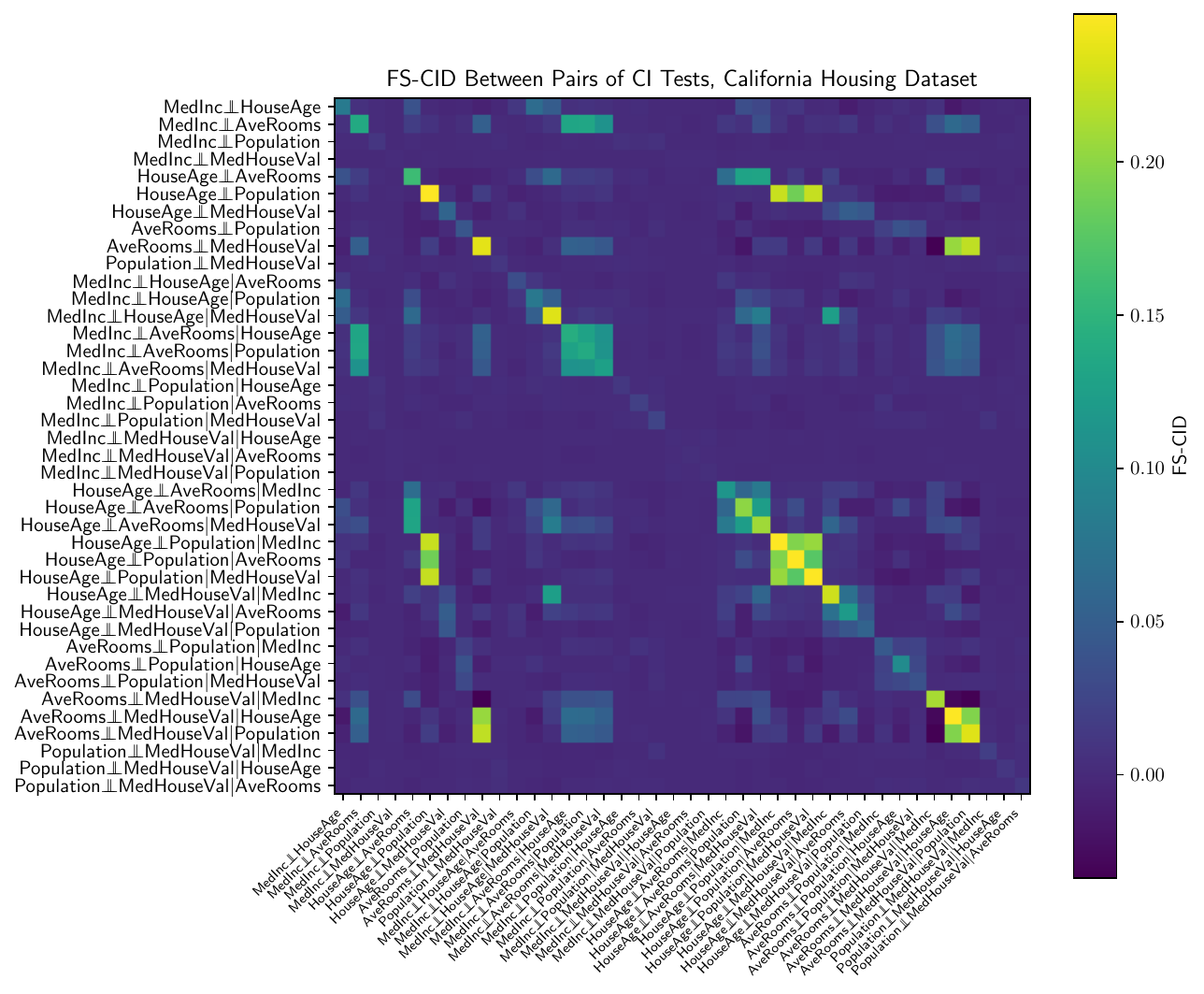}
    \includegraphics[width=0.49\linewidth]{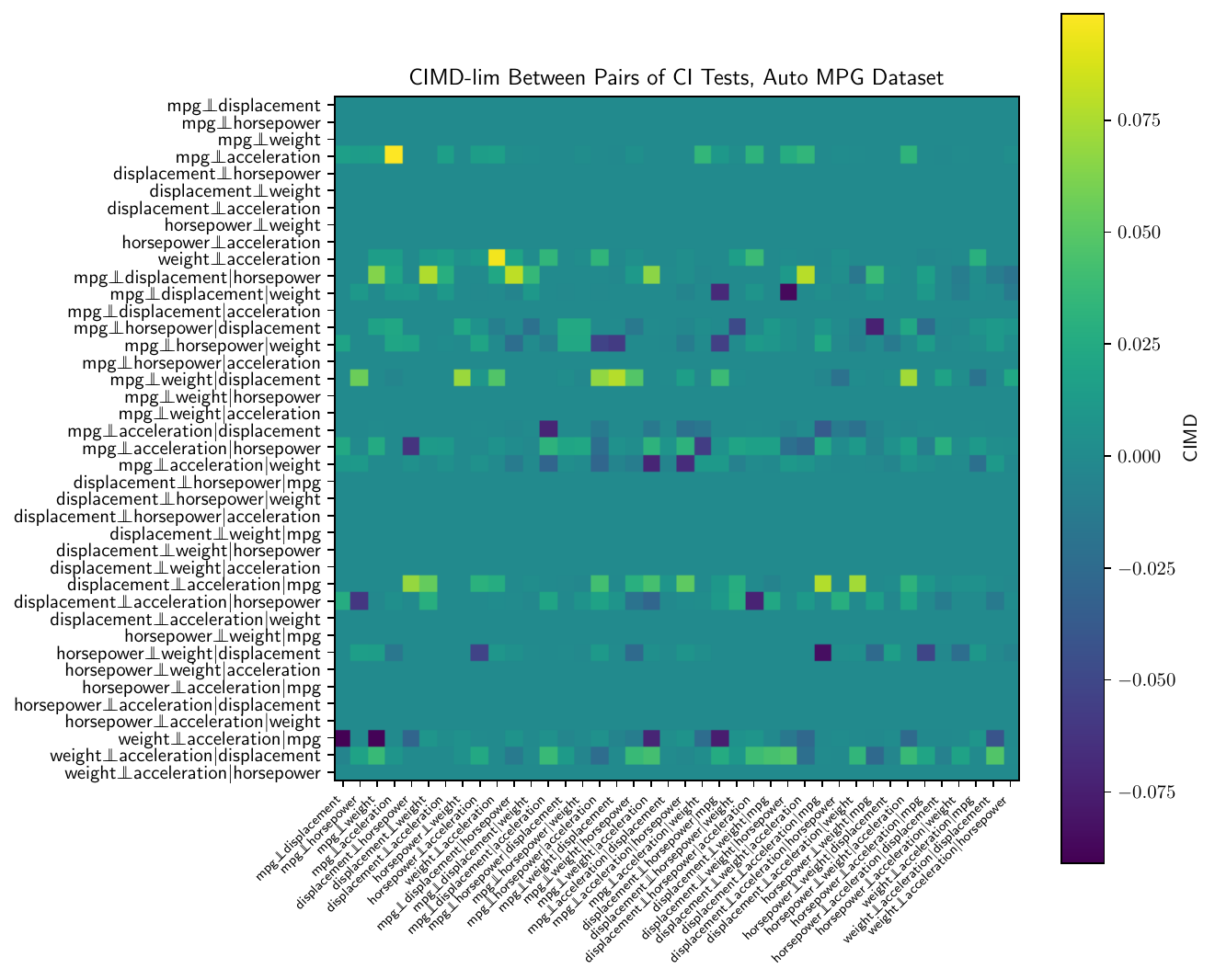}
    \includegraphics[width=0.49\linewidth]{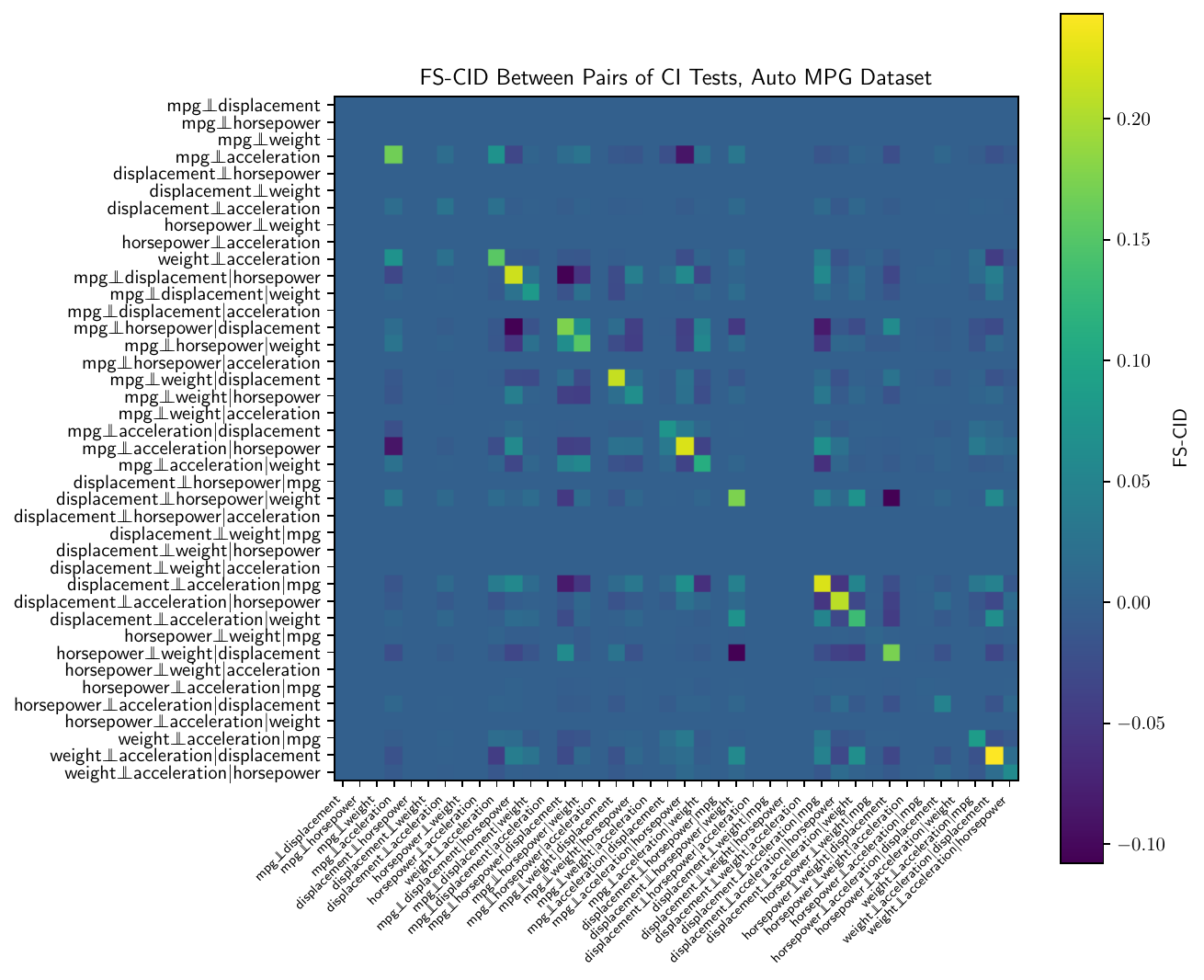}
    \caption{Demonstrating the dependence between CI-tests in real-world data using both FS-CID and limited CIMD.}
    \label{fig:real_world}
\end{figure}

Figure~\ref{fig:real_world} shows four plots on two real-world datasets. These experiments demonstrate the occurrence of conditional independence meta-dependence in real-world datasets. A shared structure is expressed between FS-CID (calculated via sub-sampling the data and computing correlation in null hypothesis rejection) and limited CIMD.

\end{document}